\documentclass[letterpaper, 10 pt, conference]{ieeeconf}  

\IEEEoverridecommandlockouts                              
\usepackage[utf8]{inputenc}
\usepackage[T1]{fontenc}
\usepackage{graphicx}
\usepackage{hyperref}
\usepackage{booktabs}
\usepackage{subcaption}
\usepackage{siunitx}
\let\labelindent\relax
\usepackage{enumitem}   
\usepackage[version=4]{mhchem}  
\usepackage{xcolor}
\hypersetup{
    colorlinks,
    linkcolor={black},
    citecolor={black},
    urlcolor={black}
}

\title{\LARGE \bf
WetExplorer: Automating Wetland Greenhouse‑Gas Surveys with an Autonomous Mobile Robot
}

\author{Jose Vasquez and Xuping Zhang$^{*}$
\thanks{Corresponding author, Department of Mechanical and Production Engineering, Aarhus University, Aarhus, Denmark.}
\thanks{
Email: xuzh@mpe.au.dk}
}
\begin{document}

\maketitle
\thispagestyle{empty}
\pagestyle{empty}

\begin{abstract}

Quantifying greenhouse‐gases (GHG) in wetlands is critical for climate modeling and restoration assessment, yet manual sampling is labor‑intensive, and time demanding. We present \textit{WetExplorer}, an autonomous tracked robot that automates the full GHG‑sampling workflow. The robot system integrates low‑ground‑pressure locomotion, centimeter‑accurate lift placement, dual‑RTK sensor fusion, obstacle avoidance planning, and deep‑learning perception in a containerized ROS2 stack. Outdoor trials verified that the sensor-fusion stack maintains a mean localization error of 1.71 cm, the vision module estimates object pose with 7 mm translational and 3° rotational accuracy, while indoor trials demonstrated that the full motion-planning pipeline positions the sampling chamber within a global tolerance of 70 mm while avoiding obstacles, all without human intervention. By eliminating the manual bottleneck, WetExplorer enables high‑frequency, multi‑site GHG measurements and opens the door for dense, long‑duration datasets in saturated wetland terrain.

\end{abstract}

\section{Introduction}
Wetlands occupy barely 1\% of the Earth's surface, but store about 20\% of all carbon in the ecosystem \cite{temmink_wetlands}.
When these systems are drained, restored or otherwise disturbed, they can emit sizable amounts of \ce{CO2}, \ce{CH4}, and \ce{N2O} \cite{petrescu2015}.
Capturing such swings requires repeated and spatially extensive measurements. This data is crucial to understand carbon cycles and its absence can limit the evaluation of costly restoration efforts. However, researchers can only obtain a few plots before labor, safety, and access constraints become a problem. Closing the gap between the scientific need for frequent sampling and the practical limitations of manual fieldwork is essential for data-driven wetland management.
\subsection{Field Conditions and Manual‑Sampling Burden}
\label{subsec:manual_burden}

\begin{figure}[!tb]
  \centering

  \includegraphics[width=1.0\linewidth]{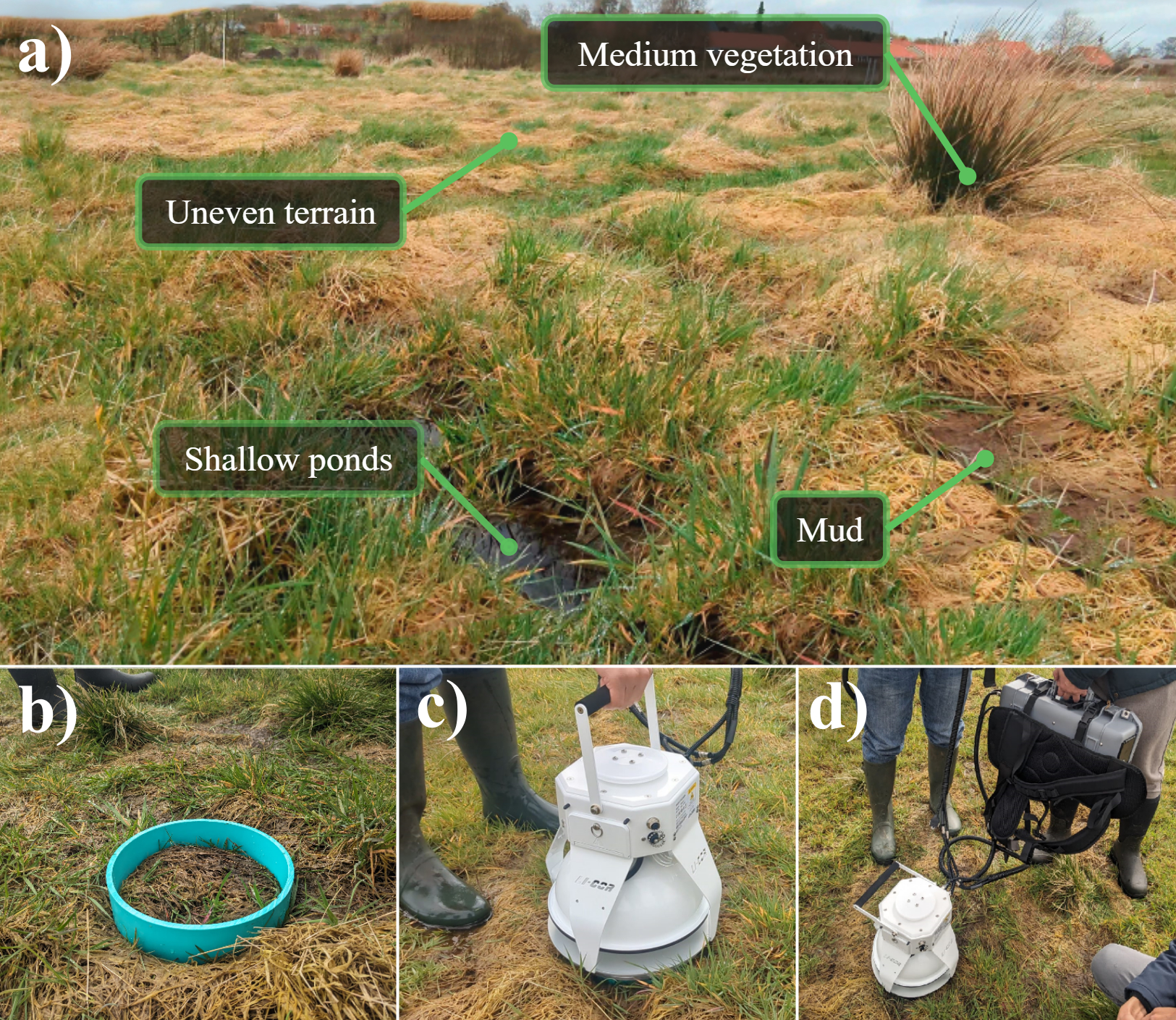}

  \captionsetup{font=small}
  \caption{Field sampling workflow in Danish wetlands. (a) Overview
 of the sampling area. (b) A PVC collar is pushed and fixed into the
 ground. (c) A chamber is seated on the collar for gas collection.
 (d) Two portable gas analyzers carried by the researchers complete the GHG readings.}
  \label{fig:manual_sampling}
\end{figure}

Our target wetlands in Denmark consist of ankle-deep mud, soft peat, knee-high vegetation, and shallow ponds (Fig.~\ref{fig:manual_sampling}a). The terrain is predominantly flat but with small vegetation clusters and peat irregularities, which typically introduce bumps of up to 10 cm in height. Carrying heavy measurement devices, such as two gas analyzers weighing \SI{10}{kg} each and a \SI{4.3}{kg} gas collection chamber, across such terrain every two hours is physically exhausting and considerably slows progress.

The current protocol (Fig.~\ref{fig:manual_sampling}b–d) requires researchers to press a PVC collar into the ground (Fig.~\ref{fig:manual_sampling}b),
then seat the chamber on the collar and then attach the tubing while holding the portable analyzer (Fig.~\ref{fig:manual_sampling}c), and
remain for the five‑minute measurement procedure before moving to the next sampling point (Fig.~\ref{fig:manual_sampling}d).
Repeating this sequence twelve times per day per collar limits most studies to only two or three plots and leaves large temporal gaps in the sampling analysis.

Automating this workflow with a robotic platform would eliminate the human bottleneck, enabling continuous, high‑resolution GHG measurements across dozens of wetlands.

\subsection{Our Contribution}

We developed WetExplorer, the first robotic system that fully automates chamber‑based greenhouse‑gas sampling in wetland terrain. Specifically, we:

\begin{enumerate}[label=(\roman*)]
  \item integrate a low‑ground‑pressure tracked chassis and design a lift mechanism capable of centimeter‑level chamber placement;
  \item implement a multisensor state‑estimation stack that fuses dual‑antenna RTK‑GPS, IMU, and encoders for robust navigation;
  \item add a mission planning layer that combines a Hybrid‑A* planner with a Pure‑Pursuit controller for global motion and a precise position controller for tool alignment; and
  \item deliver a containerized control architecture that supports unattended operation and data logging.
\end{enumerate}

Section \ref{sec:related} covers related sampling and mobile‑robot systems; Section \ref{sec:system} details the mechanical and electronic design; Section \ref{sec:navigation} explains navigation, perception, and mission software; Section \ref{sec:test} reports localization, computer vision and chamber‑placement accuracy as well a general evaluation of the system with a full‑mission field trial; and Section \ref{sec:concl} discusses limitations and future work toward greater terrain adaptability and system robustness.

\section{RELATED WORK}
\label{sec:related}

\subsection{Technologies for Outdoor Autonomous Vehicles}
\label{subsec:outdoor_nav}
Outdoor robots can localize themselves with many techniques: GNSS/RTK for global fixes, UWB or other radio beacons for local ranging, LiDAR‑ or vision‑based SLAM for map‑centric accuracy, and odometry or RFID/QR
markers for short‑range updates. A recent review shows that their performance spans 0.1m for RTK‑GNSS, centimeters for LiDAR–vision fusion (at high compute cost), and decimeters or worse for stand‑alone odometry or low‑cost radio, with each method carrying its own cost and environmental limits \cite{mobile_robot_positioning}.
Because no single sensor is fault‑proof, most field robots combine complementary sources, for example, RTK‑GPS, IMU, LiDARs, radars, depth cameras and wheel encoders to achieve the sub‑decimeter accuracy
reported on agricultural and wetland platforms.

Typical outdoor navigation couple a cost‑aware global planner with a reactive path follower.
Global planners such as grid‑based A* and Hybrid‑A* give quick, kinematically feasible routes on structured maps; sampling methods like RRT* and BIT* explore cluttered terrain more freely; and optimization or learning‑based planners, including CHOMP and value‑iteration networks, refine trajectories when vehicle dynamics or ride comfort matter \cite{wang_offroad_planning}.
Once the path is set, geometric followers like Pure Pursuit or Stanley provide lightweight steering, whereas model‑predictive and adaptive sliding‑mode controllers add dynamic awareness for tighter tracking \cite{wang_offroad_planning}.

Despite their maturity, these algorithms have not yet been validated under the slip, vegetation occlusion, and micro‑topography unique to the Danish wetlands.

\subsection{Autonomous Robots for Wetlands}
\label{subsec:wetland_robots}
Tracked or amphibious ground robots have been deployed for vegetation mapping, sediment coring, and pollution surveillance in marshes and rice paddies 
\cite{itu_amphib,azorobotics}.  
Wide tracks, paddle wheels, or oversized tires give these platforms low ground pressure, letting them traverse soft, saturated soils without sinking.  
Most payload modules, however, carry cameras or water samplers rather than mechanisms capable of making direct soil contact.

Existing aquatic robots highlight this limitation: the 5m Wivenhoe ASV carries an open‑path optical‑methane detector that “sniffs” ebullition plumes from above the water surface, while the lighter, multi‑robot Inference fleet lowers a small gas‑sampling chamber into the water for stationary flux checks \cite{boats}. Both automate methane sensing on reservoirs, but neither can press a sealed chamber against saturated soil, leaving ground‑level gas measurements unsolved for mobile platforms.

Existing systems either automate gas sampling at a single fixed point or mobilize sensors without soil contact.  
No prior work combines low‑ground‑pressure locomotion, compliant sealed placement on the target, and fully autonomous multi‑plot scheduling in wetlands, the integrated capability targeted by \textit{WetExplorer}.

\section{System Overview}
\label{sec:system}

\begin{figure*}[!tb]
  \centering
  \captionsetup{font=small}

  \begin{subfigure}[t]{0.45\textwidth}
    \centering
    \label{fig:overview}
    \includegraphics[width=\linewidth]{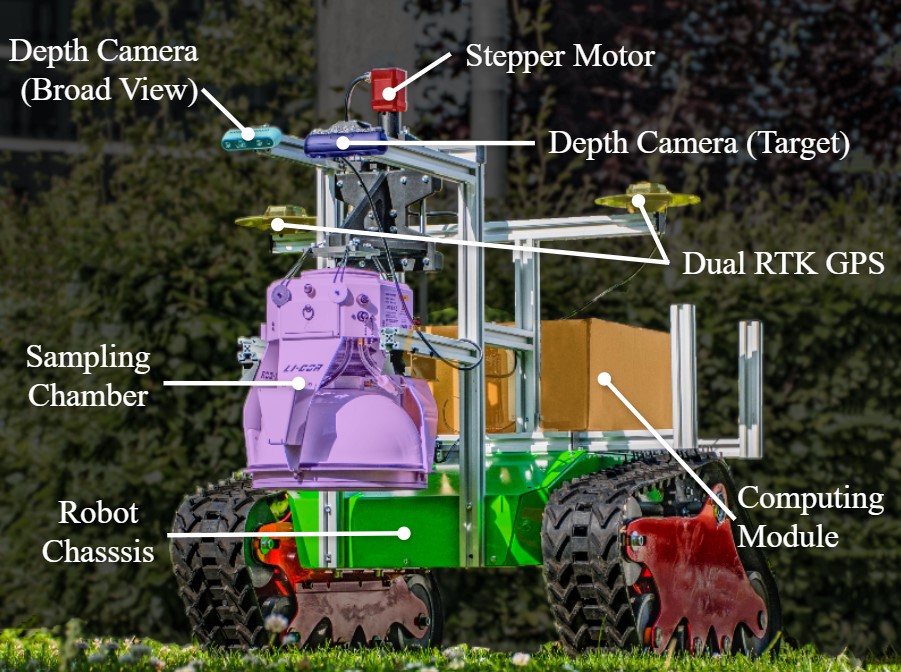}
    \small a)
    \end{subfigure}
  \hfill
  \begin{subfigure}[t]{0.54\textwidth}
    \centering
    \label{fig:arch}
    \includegraphics[width=\linewidth]{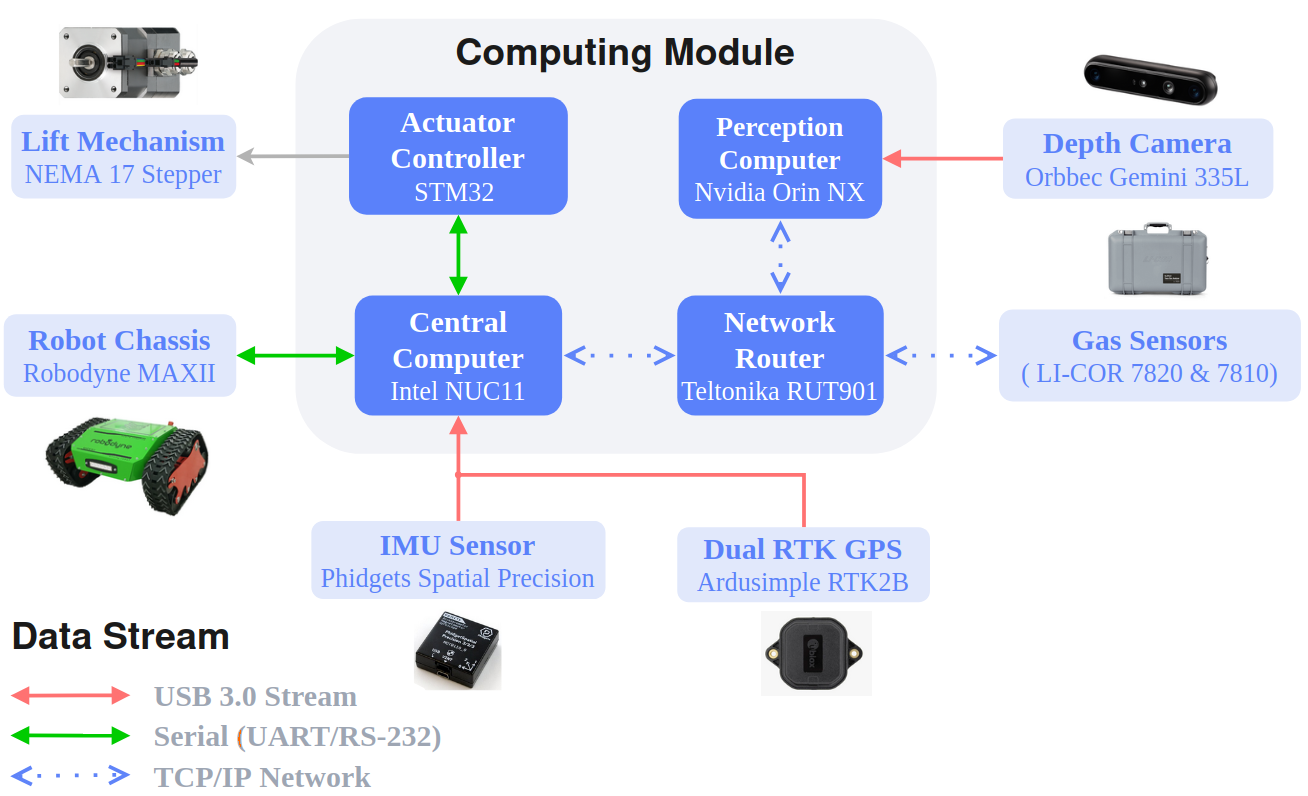}
    \small b)      
  \end{subfigure}
  \caption{\emph{WetExplorer} overview. a) Main hardware components. b) Mechatronics diagram of the computing module.}
  \label{fig:system_architecture}
\end{figure*}

Fig. \ref{fig:system_architecture}a highlights the tracked chassis capable of traversing through the rough terrain, while a screw‑ball lift precisely lowers the \SI{4.3}{kg} sampling chamber onto each collar ring.
Navigation relies on a dual‑antenna RTK‑GPS and two depth cameras: a forward‑facing, broad‑view sensor for terrain perception and a narrower target‑view sensor for pinpoint ring localization.
All sensor streams are handled by a computing module that houses the main processing and communication electronics.

Fig. \ref{fig:system_architecture}b showcases the computing stack.    
An Intel NUC serves as the central computer, interfacing with most sensors and the chassis controller, while an STM32‑based microcontroller drives the lift actuator.
An IMU and the RTK‑GPS module are fused for state estimation.
A network router links the main computation modules, including the central computer, an AI-accelerated perception computer, and the gas sensor computers, completing the modular mechatronic architecture.

\subsection{Tracked Chassis}
\label{subsec:chassis}
Wetland soils are very soft and frequently
inundated, which rules out wheeled bases and most legged platforms.
Tracked locomotion spreads the vehicle mass over a large contact area,
resulting in a low average ground pressure, well below the one produced by researchers
~\cite{track_selection}.  Among the commercial platforms surveyed, the Robodyne MAXII offered the most favorable weight-to-payload ratio (60 kg weight; 90 kg payload) and a ground clearance of 18.4 cm, making it suitable for the targeted site.

Each track is driven by a brushed DC motor coupled to a 1:30 spur gearbox. Closed-loop velocity control is managed by a Roboteq MDC2230 controller, which integrates a quadrature encoder and provides a serial interface for commanding and monitoring the chassis from an external computer. A \SI{24}{V}, \SI{35}{Ah} Li-ion battery powers the drive system for approximately \SI{4.5}{\hour} of stop-and-go operation.

\subsection{State‑Estimation Sensors}

The navigation stack fuses an inertial measurement unit (IMU) with a dual‑antenna RTK‑GPS receiver for robust, centimeter‑grade state estimation, as well as the quadrature encoders from the motors.

The chosen IMU corresponds to a PhidgetSpatial Precision 3/3/3 because it
combines a low \SI{0.0015}{\degree/s} bias gyro with a 250 Hz data rate and
on‑board AHRS computation, thereby offloading quaternion math from the
main CPU \cite{phidgetspatial}. 

In addition, the absolute position provided by the RTK-GPS is essential to revisit ring collars after multi‑hour gaps. The dual‑antenna ArduSimpleRTK2BHeading Kit provides sub‑degree heading and centimeter‑grade positioning without magnetometers \cite{ardusimple_rtk2b_heading_kit}, which can saturate in iron‑rich peat soils or be disturbed by the robot’s high‑current actuators. The module uses \SI{2.4}{GHz} radios connected to an in-house base station, delivering RTK corrections without relying on NTRIP services and maintaining reliable operation in remote areas.

\subsection{Perception Sensor}
Stereo vision enables outdoor depth perception without the range-aliasing limitations of time-of-flight sensors under sunlight. The Orbbec Gemini 335L (IP65) was selected after the Intel RealSense D435 showed good performance but lacked waterproofing. Offering similar depth quality in a sealed housing, the Orbbec meets the requirements for 3D sensing in wetlands.
\subsection{Compute Stack}
An Intel NUC11 Pro (Core i5‑1135G7, 16GB RAM)
serves as the central computer. All navigation, localization, and
mission state machines run on this host at around 30Hz.  A separate
NVIDIA Jetson Orin NX 8 GB executes learned perception tasks
(e.g., YOLOv11 ring detection, point‑cloud registration) at up to
117 TOPS, freeing the central computer for deterministic control loops and debugging.

\subsection{Networking and Telemetry}
Field sites lack Wi‑Fi, so a Teltonika RUT901
industrial LTE router provides 4G connectivity and a local network that interconnects the NUC, Jetson and Li‑Cor gas analyzers. The router runs a Wi‑Fi network that allows users to connect an external device to display the data of the robot computers.

\subsection{Lift Mechanism}
\label{subsec:lift_preview}
\begin{figure}[tb]

  \centering

  \begin{minipage}[t]{0.62\linewidth}\centering
    \includegraphics[width=\linewidth]{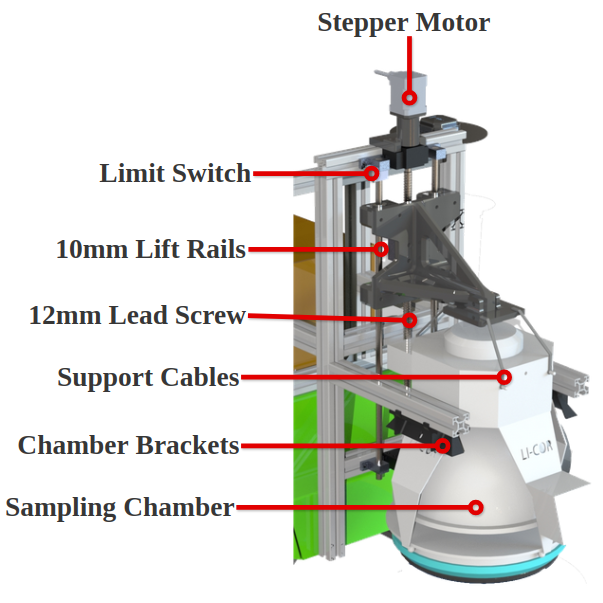}\\
    \footnotesize (a)
  \end{minipage}\hfill
  \begin{minipage}[t]{0.38\linewidth}\centering
    \includegraphics[width=\linewidth]{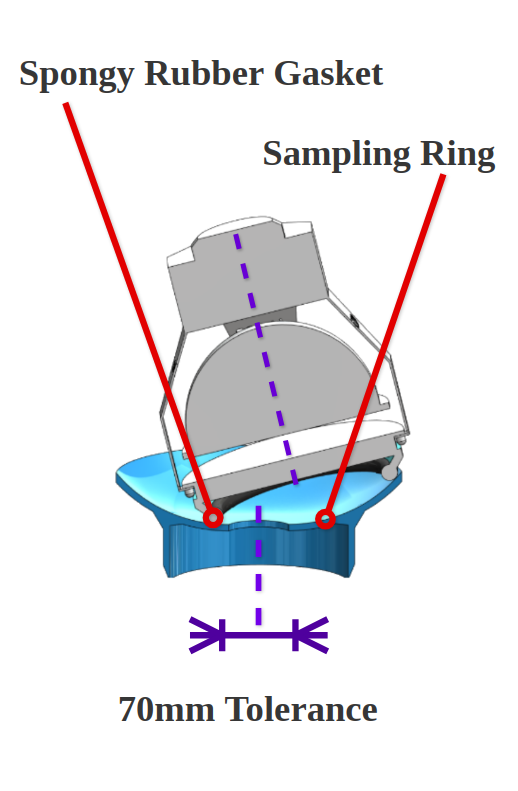}\\
    \footnotesize (b)
  \end{minipage}
\captionsetup{font=small}
\caption{Chamber Manipulator Design. a) Lift mechanism diagram. b) Section View of the Chamber - Ring coupling.}
\label{fig:lift_mechanism}
\end{figure}

The \SI{4.3}{kg} Li‑Cor chamber is lowered by a self‑locking screw–ball actuator (Fig.~\ref{fig:lift_mechanism}a); consequently, power is drawn mainly while the lift is moving.
The stepper motor runs open‑loop and re‑homes at the beginning of each cycle via a limit switch located at the top of travel.

When the robot is in motion, two locating chamber brackets lock the chamber in place to prevent swinging. Once the lift is lowered, the brackets are no longer holding the chamber and two cables act as a passive gimbal, providing a few degrees of rotational and lateral compliance as the chamber settles onto the collar.

The mechanism is driven by a Drylin® D7 stepper‑motor driver and we developed a STM32‑based controller, which converts distance commands into step pulses for the driver and manages calibration and limit-switch safety.

The collar’s spherical profile ensures a proper seal regardless of small angular misalignments, mechanically tolerating up to \SI{70}{mm} of radial placement error while compensating for minor navigation or computer vision offsets (Fig.~\ref{fig:lift_mechanism}b). A foam rubber gasket mounted on the chamber compresses under its own weight to form an airtight contact. Consequently, any successful mechanical coupling between the chamber and the collar implies that the placement error remained within this \SI{70}{mm} tolerance, which serves as the acceptance metric for evaluating the system's accuracy.

\section{AUTONOMOUS NAVIGATION AND MISSION EXECUTION}
\label{sec:navigation}
\begin{figure*}[t]
  \centering
  \includegraphics[width=0.9\textwidth]{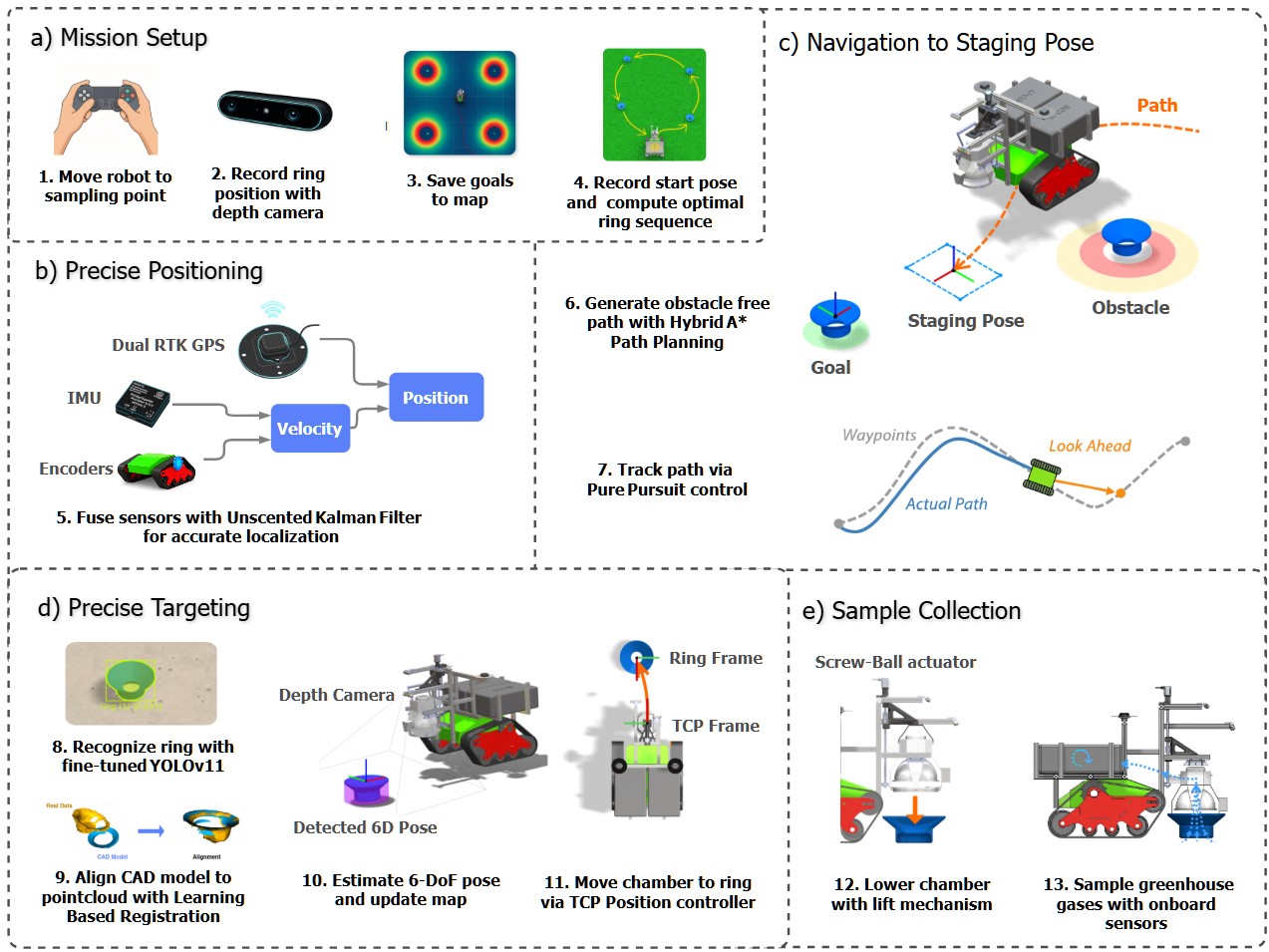}\hfill
  
  \captionsetup{font=small}
  \caption{Autonomous sampling workflow.
           a)~Mission setup: the operator drives the robot to each ring location and stores the GPS goals.  
           b)~Precise positioning: dual‑antenna RTK‑GPS, IMU, and encoders are fused by two UKFs for sub-centimeter level localization.  
           c)~Navigation to staging pose: Hybrid‑A* and Pure‑Pursuit guides the robot next to the collar while avoiding obstacles.  
           d)~Precise targeting: YOLOv11 and point‑cloud registration estimate the 6‑DoF ring pose.  
           e)~Sample collection: the lift lowers, seals, and the analyzer records \ce{CO2}, \ce{CH4}, and \ce{N2O}.}
  \label{fig:mission_workflow}
\end{figure*}

Fig. \ref{fig:mission_workflow} illustrates the end-to-end software workflow of the autonomous sampling mission. The process begins with Mission Setup (Fig.~\ref{fig:mission_workflow}a), where a map is generated and the required sampling points are defined. Precise Positioning (Fig.~\ref{fig:mission_workflow}b) establishes a closed state-estimation loop to maintain accurate localization. During Navigation to Staging Pose (Fig.~\ref{fig:mission_workflow}c), the robot autonomously travels to the vicinity of each collar. Precise Targeting (Fig.~\ref{fig:mission_workflow}d) refines the detected ring pose and guides the sampling chamber into alignment. Finally, Sample Collection (Fig.~\ref{fig:mission_workflow}e) lowers the chamber and records gas measurement data. The following section details the algorithms and implementation strategies that enable this fully unattended operation.

\subsection{Kinematic Model and Odometry}
\label{subsec:kinematics}
The tracked base is modeled as a differential‑drive platform with
equivalent sprocket radius~$r$ and track separation ~$b$.  
The inverse‑kinematics equations convert the commanded linear and
angular velocities~$\bigl(V_{x},\dot{\theta}\bigr)$ into left and right
sprocket angular speeds~$(\omega_{L},\omega_{R})$:

\begin{equation}
\begin{aligned}
  \omega_{L} &= \frac{V_{x}-\tfrac{1}{2}\dot{\theta}\,b}{r},\quad
  \omega_{R} &= \frac{V_{x}+\tfrac{1}{2}\dot{\theta}\,b}{r}.
\end{aligned}
\end{equation}

On the other hand, the forward kinematics equations convert the measured motor
speeds (reported in r/min) to rad/s and reconstruct the robot’s twist:

\begin{equation}
\begin{aligned}
  V_{x}= \frac{v_{L}+v_{R}}{2}, \qquad
  \dot{\theta}= \frac{v_{R}-v_{L}}{b}.
  \end{aligned}
\end{equation}

where $v_{L}$ and $v_{R}$ are the left and right linear track speeds, respectively.

The resulting odometry stream is fused with inertial and RTK‑GPS data,
as described next.

\subsection{Localization through Sensor Fusion}
Using a single sensor is insufficient in wetlands: IMU integration drifts, wheel odometry slips, and low‑rate GPS misses fast maneuvers. We therefore adopt a loosely coupled Unscented Kalman Filter (UKF) to fuse data from all sensors, employing the widely used and robust \textit{robot\_localization} implementation in ROS2 \cite{ekf}.

\subsubsection*{Velocity Sensor Fusion}
combines IMU accelerations and angular rates with odometry, using manufacturer-based IMU covariances and adaptive odometry noise that increases with speed to account for slip \cite{slippage}.

\subsubsection*{Position Sensor Fusion}
augments this state with dual-antenna RTK-GPS fixes at 5 Hz. The GPS position ($\sigma\!\approx\!1.5$cm)  dominates the update, while velocity constraints smooth the trajectory and eliminate jumps in raw GPS data, producing a reliable 20 Hz localization stream for navigation.

\subsection{Navigation to the Staging Pose}
Mission way‑points (ring collars) are allocated in
Fig.~\ref{fig:mission_workflow}c.  Each transfer employs:

\paragraph*{Global planner}
Smac Hybrid‑A* is selected because it produces
kinematically feasible paths for non‑holonomic vehicles while still
replanning fast enough for receding‑horizon use on embedded CPUs
\cite{open_kin_planner}
.  
The solver was configured for our tracked base by imposing a minimum turn radius of 0.20 m, a cost penalty of 3.0, a reverse penalty of 2.0, a non-straight penalty of 1.0, and a change penalty of 5.0. These parameters penalize zig-zag motions, yielding smoother trajectories that are easier for the controller to track.

Wetland “no‑go” areas (vegetation, water ponds, neighboring collars) 
are converted into terrain masks and inflated on the custom costmap; the higher
cell costs, together with the penalties above, steer the solution away
from surfaces that could lead to collisions or sink‑ins.

\paragraph*{Local controller}
Regulated Pure Pursuit (RPP) is adopted instead of the standard
pure‑pursuit tracker because it keeps the same geometric steering but
adds built‑in speed regulation for tight corners and crowding obstacles,
a decisive advantage for our heavy, slip‑prone tracked base
\cite{regulated_pure_pursuit}.  
The controller follows the Hybrid‑A* spline while continuously lowering
the commanded speed when curvature exceeds a set radius or obstacles enter a safety band, two effects that matter on the cluttered service routes of our test site.

The principal controller parameters, identified through simulation, are a \textit{lookahead time} of 1.5\,s, a \textit{desired linear speed} of 0.20\,m/s, and a \textit{rotate-to-heading} protocol that engages whenever the robot’s orientation deviates by at least 45$^{\circ}$ from the path.  At nominal cruise speed these settings yield a \textit{lookahead distance} of roughly 0.6\,m, which is equal to the 0.6\,m wheelbase.

\subsection{Ring 6‑DoF Pose Estimation}
Our ring-pose estimation pipeline is carried out by a three-stage pipeline that remains light enough for online execution on the Jetson Orin NX. First, a fine-tuned \emph{YOLOv11-n-seg} network detects the ring at 15 fps and returns a binary mask, offering a good speed–accuracy trade-off on edge GPUs \cite{yolo}. The mask is then applied to the depth image so that only valid 3D points are retained; then a centroid computation supplies a fallback pose whenever the subsequent alignment step is unsuccessful. Finally, the cropped point-cloud is aligned to a CAD template that holds the reference of the object with Predator, a learning-based point-cloud matcher tolerant of low surface overlap \cite{predator}. A single ICP refinement reduces the error and increases the final overlap, delivering a camera-to-ring transform that easily satisfies the robot’s positioning tolerance \cite{handeye}.

The target object is placed in an open area where it is rarely occluded by surrounding elements, and its orientation remains relatively stable under different circumstances. Varying illumination conditions, including low light, direct sunlight, and shadowed areas, were considered during dataset preparation. The object detection module was trained on 100 positive and 100 negative RGB samples, which were augmented using brightness variation, Gaussian blur, and noise injection, resulting in a dataset of 600 images that combined indoor and outdoor scenes under both sunny and cloudy conditions. For the point-cloud alignment stage, 50 primarily outdoor point-cloud pairs were collected and augmented to 250 pairs using random spatial transformations, added noise, and simulated partial observations to improve robustness in real-world deployments.

\subsection{Chamber Alignment}
The tool center point (TCP) is guided towards the ring by a
planar PI controller (step\,11,
Fig.~\ref{fig:mission_workflow}c). The control loop is inspired by a classic mobile robot feedback control while leaving the final orientation unconstrained \cite{mobile_robots}.  Let
$\mathbf{p}_{\text{ring}}$ and $\mathbf{p}_{\text{TCP}}$
be the positions of the ring and the TCP in the map frame.
The in‑plane error vector is
\[
\begin{aligned}
\boldsymbol{\Delta p} &= (\Delta x,\;\Delta y)^{\mathsf T}  &=  \mathbf{p}_{\text{ring}}
                         - \mathbf{p}_{\text{TCP}} .
\end{aligned}
\]

\paragraph{Error signals}
\begin{gather}
e_p = \sqrt{\Delta x^{2} + \Delta y^{2}}, \\
e_\alpha= \operatorname{atan2}(\Delta y,\Delta x)
          \;-\;\psi ,
\end{gather}
where $\psi$ is the current heading angle of the robot.

\paragraph{Control law}
\begin{align}
v      &= K_{P_p}\,e_p
         + K_{I_p}\!\int_0^{t} e_p(\tau)\,\mathrm d\tau, \\[2pt]
\omega &= K_{P_\alpha}\,e_\alpha
         + K_{I_\alpha}\!\int_0^{t} e_\alpha(\tau)\,\mathrm d\tau ,
\end{align}
with $v$ the linear and $\omega$ the angular velocity commands.
These speeds are finally transformed from the TCP frame into the robot base frame before being forwarded to the low‑level motion controller.
\paragraph{Tuning}
The proportional and integral gains
$(K_{P_p},K_{I_p},K_{P_\alpha},K_{I_\alpha})$
were first tuned in the Gazebo simulation and then adjusted on the
real robot to compensate for inaccuracies of the models.

\subsection{Digital‑Twin Simulation}
All navigation software was first validated in Gazebo Harmonic.  A xacro-based robot
model includes mass/inertia for each link; sensor plug‑ins
replicate IMU drift, GPS precision, and depth‑cameras, enabling hardware‑in‑the‑loop tests before field deployment. 
The tracks are simulated using a kinematic approach that computes the ground forces exerted by the track surfaces to achieve the desired velocity. Instead of simulating the tracks' physical movement along pulleys, this method applies virtual forces that push the robot. This approach is computationally efficient and delivers accurate results, making it the most suitable implementation in Gazebo for track-based robots~\cite{tracks}.

\subsection{Software Implementation}
\label{subsec:sw_arch}
\begin{figure}[tb]
\centering
\includegraphics[width=1.0\linewidth]{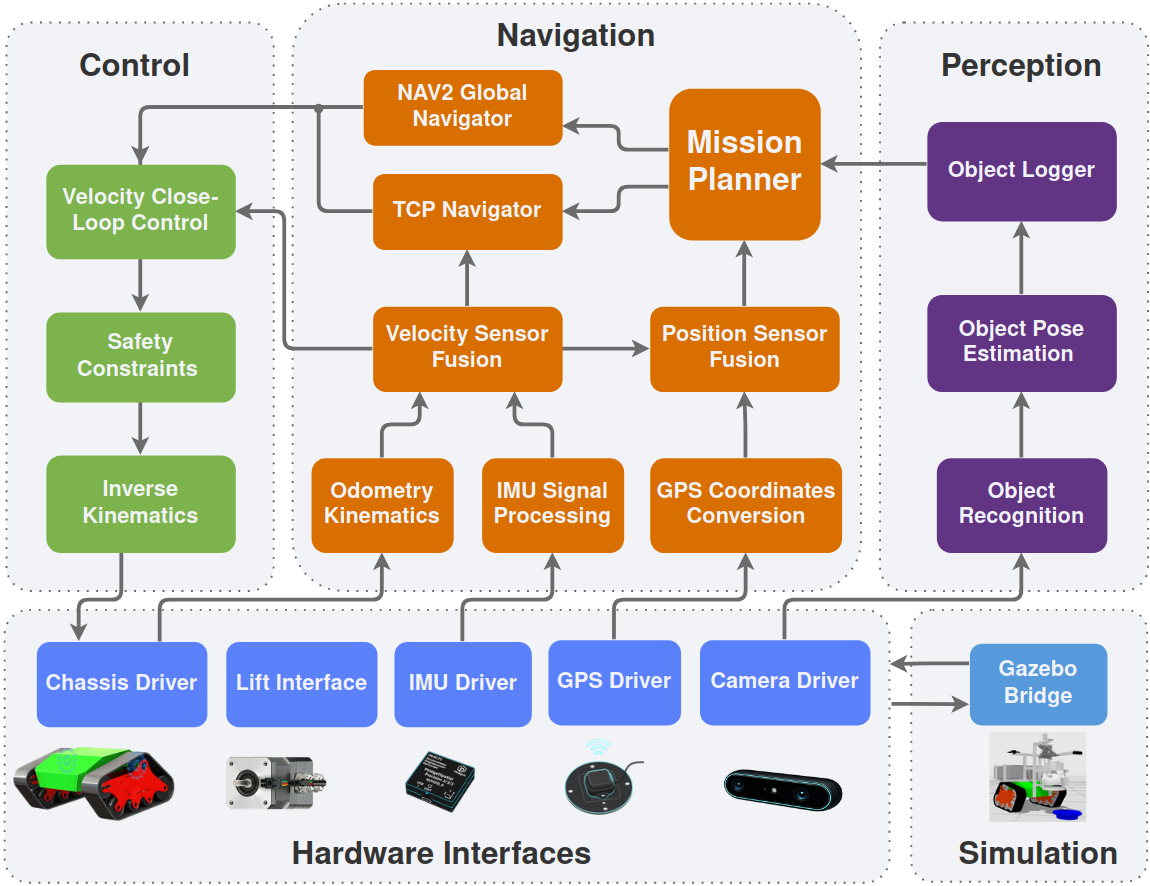}
\captionsetup{font=small}
\caption{ROS2 Software Architecture}
\label{fig:software_architecture}
\end{figure}

The robot’s software stack is built on ROS2 for its modular architecture, visualization tools and sensor drivers available. As illustrated in Fig.  \ref{fig:software_architecture}, the nodes are divided into four logical layers: a hardware‑interface layer that handles the motor driver, IMU, and RTK receiver; a control layer implementing inverse and forward kinematics together with PID velocity loops; a navigation layer responsible for sensor fusion, cost‑map planning, and path planning; and a perception layer that combines YOLOv11 object detection with point‑cloud alignment.
All workflows are packaged as
Docker images, enabling one‑command deployment and reproducible builds.

The complete source code is available as an
open‑source repository at \cite{WetExplorerRepo}.

\section{Experimental Validation}
\label{sec:test}
\subsection{Localization Benchmark}
\label{subsec:loc_bench}
\begin{figure}[t]
  \centering
  \includegraphics[width=\linewidth]{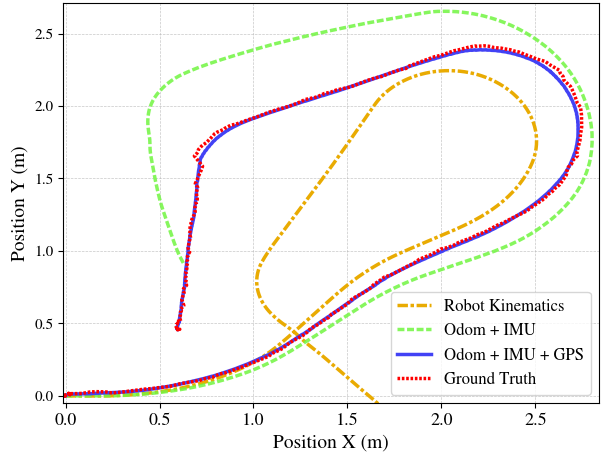}
  \captionsetup{font=small}
  \caption{Planar (X–Y) trajectories recorded during the localization benchmark. 
  Yellow: wheel‐encoder odometry that accumulates drift over time; 
  blue: velocity estimate obtained by fusing IMU and encoder data; 
  green: position estimate produced by integrating the fused velocity and GPS fixes; 
  red: high‑accuracy ground truth from the RTK‑GPS receiver.}
  \label{fig:loc_benchmark}
\end{figure}

The localization tests were conducted outdoors under good GPS coverage, similar to the open-field conditions expected at the deployment sites. The robot operated on a grassy area following a closed-loop trajectory of approximately \SI{7}{\metre} at a constant speed of \SI{0.5}{\metre\per\second}. Four pose estimates were logged simultaneously (Fig.~\ref{fig:loc_benchmark}). The resulting 2D trajectories, shown in the first graph, provide an overview of the motion smoothness and the performance differences among the methods. To complement this, Table~\ref{tab:localization_accuracy} compares the 3D position estimates against the ground truth and summarizes the mean errors for each method.

Centimeter-level \emph{RTK-GPS} consists in the ground truth, revealing the magnitude and nature of the errors in the other localization outputs. The first estimate, generated by pure \emph{forward kinematics} (yellow trace), integrates only wheel-encoder ticks and therefore exhibits cumulative drift caused by the uncertainty in the differential-drive model’s angular-speed estimation. The second estimate, the \emph{Velocity UKF} (green trace), fuses encoder-derived wheel speeds with the \SI{250}{Hz} IMU using a velocity-space unscented Kalman filter. Although this approach effectively captures yaw rates and mitigates short-term slip, it cannot compensate for lateral drift during turns, resulting in long-term trajectory deviation. Finally, the \emph{Position UKF} (blue trace) fuses RTK-GPS, IMU, and odometry to produce the navigation solution used in the field. This estimate closely follows the ground truth while maintaining the smooth motion of the Velocity UKF, effectively eliminating the jumps inherent in raw GPS data. 

The experiment was repeated across three outdoor trajectories, and the resulting errors are summarized in Table~\ref{tab:localization_accuracy}. Forward kinematics and velocity-space fusion methods accumulated larger errors over time, yet the results clearly show that the sensor fusion stage improved both heading and position accuracy, reducing errors from \SI{0.5}{\metre} to \SI{0.17}{\metre} and from \SI{8.3}{\degree} to \SI{2.6}{\degree}. The final fused estimate achieved an average 3D position error of \SI{1.71}{\centi\metre} and a heading error of \SI{0.18}{\degree}. These values fall within the precision reported by the RTK-GPS (\SI{2.36}{\centi\metre}, \SI{0.94}{\degree}), confirming that the fusion model matches GPS accuracy while providing smoother and more consistent motion estimates.
This shows that position is effectively estimated and that the navigation stack benefits from fast and accurate localization.

\begin{table}[!t]
\centering
\caption{Localization Accuracy Across Sensor Fusion Levels}
\setlength{\tabcolsep}{3pt}
\renewcommand{\arraystretch}{1.0}
\begin{tabular}{lcccc}
\toprule
\textbf{Method} & 
\textbf{Pos.\ Err. (m)} & 
\textbf{Std. (m)} & 
\textbf{Head.\ Err. ($^\circ$)} & 
\textbf{Std. ($^\circ$)} \\
\midrule
\textit{Ground Truth}     & - & 0.0236 & - & 0.954 \\
\textit{Kinematics}       & 0.509 & 0.403  & 8.27  & 6.12  \\
\textit{Odom + IMU}       & 0.171 & 0.113  & 2.61  & 2.59  \\
\textit{Odom + IMU + GPS} & 0.017 & 0.011  & 0.19  & 0.54  \\
\bottomrule
\end{tabular}
\label{tab:localization_accuracy}
\end{table}

\subsection{Computer Vision Validation}
Validation focused on the registration stage, as it most directly determines pose accuracy. To thoroughly assess robustness, the experiments were designed to include challenging scenarios, such as the target object being partially occluded by the gas chamber or cropped by the camera’s field-of-view limits. Multiple observations of the object were collected at various positions within the camera’s view under both indoor and outdoor lighting conditions.

\begin{figure}[tb]
  \centering
  \includegraphics[width=\linewidth]{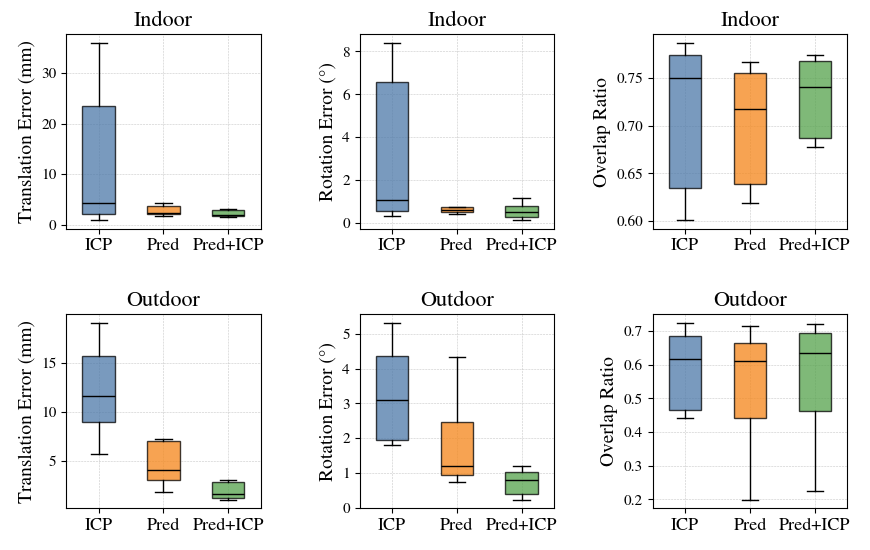}\\
  \captionsetup{font=small}  
  \caption{Box plots showing the evaluation of the Object Localization in the indoor and outdoor testing sites.} 
  \label{fig:indoor_eval}
\end{figure}

Each algorithm was evaluated on ten indoor and ten outdoor setups, executed ten times per setup. Fig. \ref{fig:indoor_eval} compares ICP, Overlap Predator, and Predator + ICP, showing that Predator significantly reduces variability in translation and rotation, while the final ICP refinement further improves precision and overlap.

The mean values corresponding to these plots are summarized in Table~\ref{tab:registration_performance}. Under indoor conditions, standard ICP achieved an average translational error of \SI{8.16}{\milli\metre}, whereas the proposed method reduced it to \SI{1.53}{\milli\metre}, also yielding a higher overlap ratio. Outdoor experiments proved more challenging due to lower point-cloud quality; however, the proposed approach maintained robust performance, achieving a mean positional error of \SI{7.19}{\milli\metre} and an orientation error of \SI{3.17}{\degree}. It effectively doubled the accuracy of standard ICP and achieved a 100\% success rate with at least 0.2 overlap, compared to 90\% for ICP.

These results confirm that the proposed method provides sub-centimeter object-localization accuracy and remains reliable under outdoor and partially occluded conditions, meeting the precision requirements for accurate ring positioning on the robot.

\begin{table}[!t]
\centering
\caption{Registration Performance Across Environments}
\setlength{\tabcolsep}{2.8pt} 
\renewcommand{\arraystretch}{0.95} 
\begin{tabular}{lccccc}
\toprule
\textbf{Env.} & 
\textbf{Method} & 
\textbf{Pos. (mm)} & 
\textbf{Rot. ($^\circ$)} & 
\textbf{Overlap} & 
\textbf{Succ. (\%)} \\
\midrule
  & ICP         & 8.16  & 4.27  & 0.69 & 90 \\
Indoor& Predator    & 2.69  & 0.98  & 0.70 & 100 \\
        & Pred + ICP  & 1.53  & 0.60  & 0.73 & 100 \\
\midrule
 & ICP         & 16.33 & 6.32  & 0.48 & 90 \\
Outdoor& Predator    & 9.08  & 3.04  & 0.47 & 90 \\
        & Pred + ICP  & 7.19  & 3.17  & 0.49 & 100 \\
\bottomrule
\end{tabular}
\label{tab:registration_performance}
\end{table}

\subsection{Demonstration: Multi‑Ring Mission Indoors}
\label{subsec:demo}

\begin{figure}[tb]
  \centering

  \begin{minipage}[t]{0.499\linewidth}\centering
    \includegraphics[width=\linewidth]{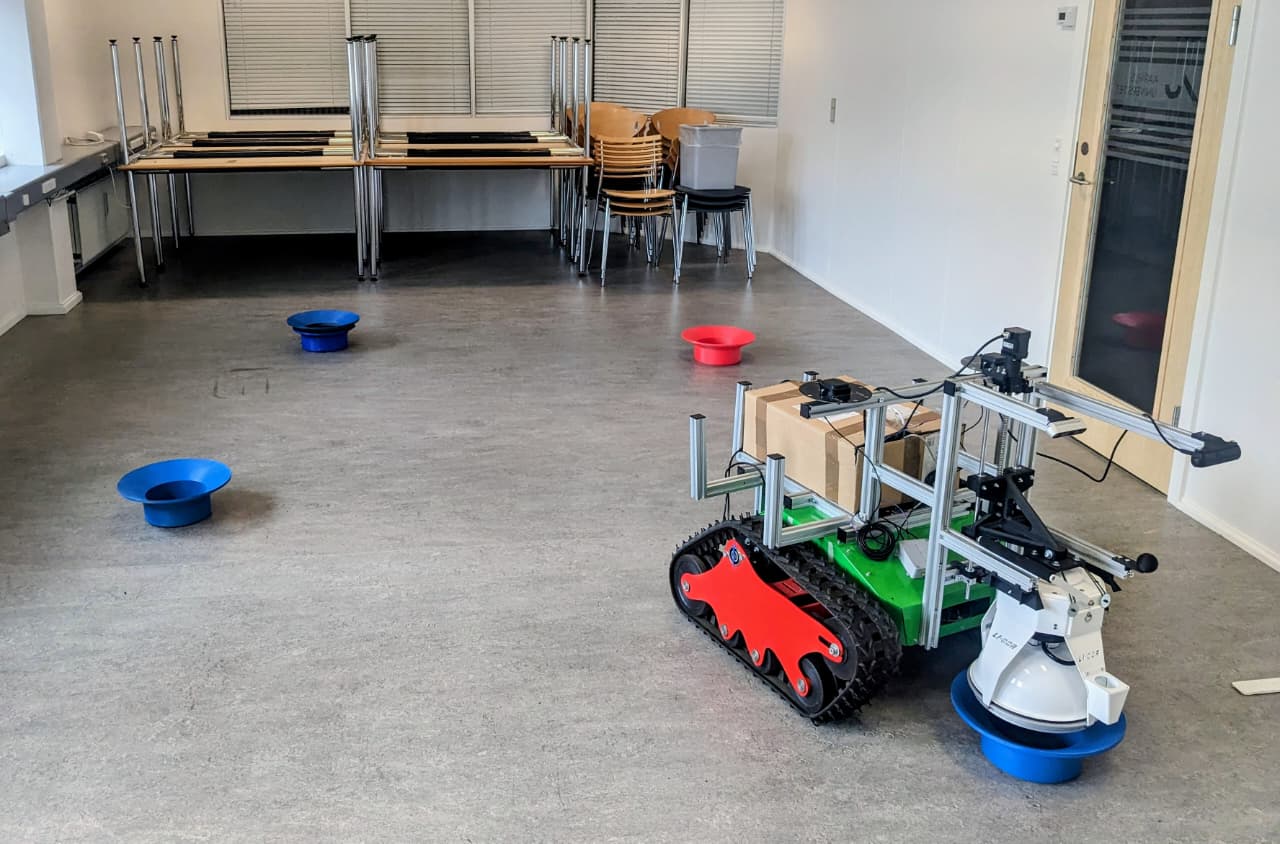}\\
    \footnotesize (a)
  \end{minipage}\hfill
  \begin{minipage}[t]{0.499\linewidth}\centering
    \includegraphics[width=\linewidth]{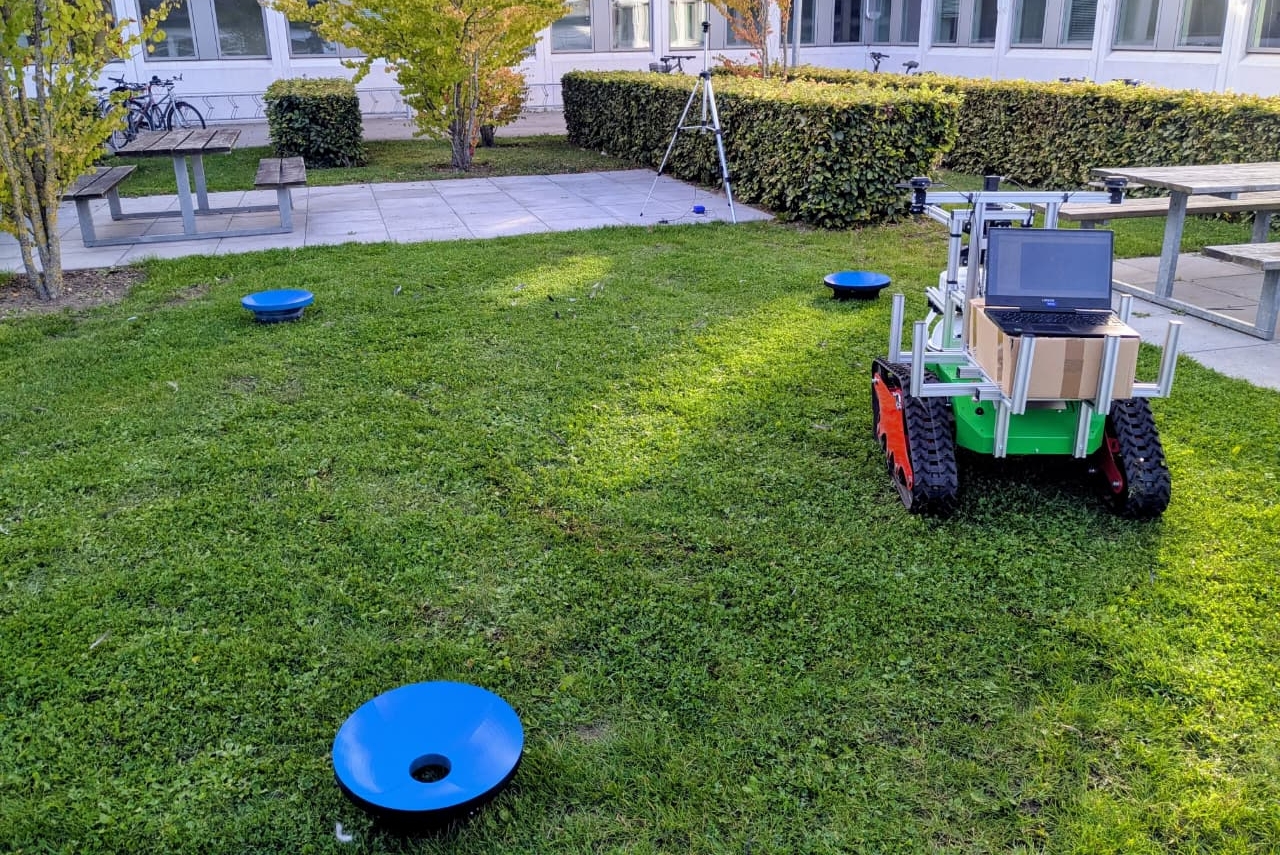}\\
    \footnotesize (b)
  \end{minipage}




  \caption{Multi‑ring demonstration testing zones. a) Indoor Testing Area. b) Outdoor Testing Area.}
  \label{fig:demo_photos}
\end{figure}

The navigation algorithms were tested in two sites: an indoor environment for algorithm tuning and an outdoor area for system validation under realistic conditions (Fig.~\ref{fig:demo_photos}).
Navigation was divided into two stages: (i) reaching the staging pose and (ii) accurately placing the sampling chamber on the ring. In total, 12 indoor and 6 outdoor missions were performed across a 5m~$\times$~5m grid.

The lack of GPS of indoors led to localization drift and reduced obstacle-avoidance reliability (91.7\%). The robot occasionally collided with obstacles due to IMU heading drift and kinematic model simplifications that assume ideal rotations. Mapping errors also caused staging poses to deviate from the target rings. A recovery routine using computer-vision feedback was added, allowing the robot to correct its pose and ensure reliable TCP placement. The average time to reach the staging pose was 14.2 s (Table~\ref{tab:nav_validation}), indicating efficient path planning. Since sampling missions occur roughly every two hours, time performance was not a limiting factor.

Outdoor tests were conducted on grass terrain at the university. Although less demanding than wetlands, they revealed several design challenges. With stable GPS reception, localization improved, but the robot initially struggled with slopes and chamber placement; the hanging chamber changes its final position according to the orientation, affecting accuracy. Localization was therefore extended to 3D, and the TCP placement algorithm was modified to predict contact based on gravity. Slippage and traction issues were mitigated by adding a velocity control loop to enforce motion commands. After these adjustments, the system achieved consistent obstacle avoidance and goal-reaching performance (Table~\ref{tab:nav_validation}), with slightly longer travel times due to rougher terrain.

TCP placement used only odometry and IMU data, as GPS noise caused small jumps. The short 45 cm distance between staging pose and target minimized drift. Indoor tests achieved 5.5 mm placement accuracy and outdoor trials 12.7 mm, both well within the 70 mm tolerance required for a sealed coupling. The higher outdoor variation reflected terrain irregularities.

These experiments validated the integration of localization, perception, and navigation, showing that missions can be executed autonomously and repeatedly in realistic environments. The results also establish a solid baseline for evaluating and comparing future navigation and perception methods under similar terrain conditions. Remaining challenges such as slippage, soft ground, standing water, and reflections will be addressed in future iterations to achieve fully autonomous operation in natural wetlands.

\begin{table}[!t]
\centering
\caption{Navigation and Placement Validation Targets}
\setlength{\tabcolsep}{2.5pt} 
\renewcommand{\arraystretch}{1.05} 
\begin{tabular}{lcc}
\toprule
\textbf{Metric} & \textbf{Indoor} & \textbf{Outdoor} \\
\midrule
Obstacle Avoidance Success & $91.7\%$ & $100\%$ \\
Average Time to Staging Pose & $14.2 s$ & $18.6s$ \\
TCP Placement Accuracy & $5.51$ mm & $12.67$ mm \\
TCP Repeatability~$\sigma$ & $2.40$\,mm & $4.11$\,mm \\
\bottomrule
\end{tabular}
\label{tab:nav_validation}
\end{table}

\section{Conclusion and Future Work}
\label{sec:concl}

\textbf{Conclusion—}This paper has presented the development and testing of \textit{WetExplorer}, a fully integrated ROS~2 robotic platform that unifies perception, navigation, and control. The design accommodates the specialized sensors required for wetland gas sampling, delivers robust traversability, and positions the collection chamber within a 70\,mm tolerance. Outdoor experiments confirmed that the sensor‐fusion stack achieves centimeter-level localization accuracy with smoothly updated estimates at 20\,Hz. Indoor trials validated the vision pipeline, yielding an average positional error of 7\,mm and an orientation error of 3.2$^{\circ}$ while operating in real time. Navigation mission indoor and outdoor testing demonstrated consistent reliability: the system planned collision-free paths and placed the chamber on the target rings well within the 70\,mm tolerance, effectively transforming a labor-intensive manual protocol into a single-command robotic operation.

\textbf{Future Work—}We will extend the system’s robustness in harsher, wetland‑like conditions by  
(i) mechanically hardening the chassis and lift to tolerate deeper inundation and floating debris;  
(ii) adapting the control architecture to the specific terrain characteristics of the field site;   
(iii) augmenting the software stack with a terrain‑traversability analysis to detect sinkholes and obstructing vegetation; and  
(iv) validating long‑duration missions across larger, heterogeneous peatland sites.  
These upgrades will advance \textit{WetExplorer} from controlled test plots to fully autonomous, high‑resolution greenhouse‑gas monitoring in real‑world wetlands.

\section*{Acknowledgment}
The authors would like to thank Dr. Shubiao Wu and his team for their assistance on wetland sampling, and the Carlsberg Foundation for the financial support that made this project possible.

\addtolength{\textheight}{-12cm}   





\bibliographystyle{IEEEtran}   
\bibliography{references}         

@article{petrescu2015,
  author  = {Petrescu, A. M. R. and Lohila, A. and Tuovinen, J.-P. and Baldocchi, D. D. and
             Desai, A. R. and Roulet, N. T. and Vesala, T. and Dolman, A. J. and
             Oechel, W. C. and Marcolla, B.},
  title   = {The uncertain climate footprint of wetlands under human pressure},
  journal = {Proceedings of the National Academy of Sciences},
  volume  = {112},
  number  = {15},
  pages   = {4594--4599},
  year    = {2015},
  doi     = {10.1073/pnas.1416267112}
}

@misc{itu_amphib,
  author       = {{ITU} {AI for Good}},
  title        = {From Research to Rescue: Amphibious Robots Transforming Environmental Monitoring},
  year         = {2024},
  howpublished = {\url{https://aiforgood.itu.int/from-research-to-rescue-amphibious-robots}},
  note         = {Accessed 2025-07-21}
}

@article{azorobotics,
  author  = {Cohen, A. and Zarrouk, D.},
  title   = {Design, Analysis and Experiments of a High-Speed Water Hovering Amphibious Robot: AmphiSTAR},
  journal = {IEEE Access},
  pages   = {1--1},
  year    = {2023},
  doi     = {10.1109/ACCESS.2023.3299498}
}

@misc{WetExplorerRepo,
  author       = {V\'azquez-Rojas, J. P.},
  title        = {{WetExplorer--Autonomous Tracked Robot} (main)},
  year         = {2025},
  howpublished = {\url{https://github.com/josepablovr/WetExplorer-Autonomous-Tracked-Robot}},
  note         = {Accessed 2025-07-21}
}

@article{track_selection,
  author  = {Bruzzone, L. and Quaglia, G.},
  title   = {Locomotion Systems for Ground Mobile Robots in Unstructured Environments},
  journal = {Mechanical Sciences},
  volume  = {3},
  pages   = {49--62},
  year    = {2012},
  month   = jul,
  doi     = {10.5194/ms-3-49-2012}
}

@article{tracks,
  author         = {Pecka, M. and Zimmermann, K. and Svoboda, T.},
  title          = {Fast Simulation of Vehicles with Non-Deformable Tracks},
  journal        = {arXiv preprint},
  archivePrefix  = {arXiv},
  eprint         = {1703.04316},
  year           = {2017},
  month          = mar,
  doi            = {10.48550/arXiv.1703.04316}
}

@article{handeye,
  author  = {Li, L. and Yang, X. and Wang, R. and others},
  title   = {Automatic Robot Hand--Eye Calibration Enabled by Learning-Based 3D Vision},
  journal = {Journal of Intelligent \& Robotic Systems},
  volume  = {110},
  pages   = {130},
  year    = {2024},
  doi     = {10.1007/s10846-024-02166-4}
}

@article{yolo,
  author        = {Khanam, R. and Hussain, M.},
  title         = {YOLOv11: An Overview of the Key Architectural Enhancements},
  journal       = {arXiv preprint},
  archivePrefix = {arXiv},
  eprint        = {2410.17725},
  year          = {2024},
  url           = {https://arxiv.org/abs/2410.17725}
}

@article{predator,
  author        = {Huang, S. and Gojcic, Z. and Usvyatsov, M. and Wieser, A. and Schindler, K.},
  title         = {PREDATOR: Registration of 3D Point Clouds with Low Overlap},
  journal       = {arXiv preprint},
  archivePrefix = {arXiv},
  eprint        = {2011.13005},
  year          = {2021},
  url           = {https://arxiv.org/abs/2011.13005}
}

@book{mobile_robots,
  author    = {Siegwart, R. and Nourbakhsh, I.},
  title     = {Introduction to Autonomous Mobile Robots},
  series    = {Intelligent Robotics and Autonomous Agents},
  address   = {Cambridge, MA},
  publisher = {MIT Press},
  year      = {2004},
  isbn      = {0-262-19502-X},
  note      = {``Mobile Robot Kinematics,'' pp.\,83--88}
}

@article{regulated_pure_pursuit,
  author  = {Macenski, S. and Singh, S. and Mart\'in, F. and Gin\'es, J.},
  title   = {Regulated Pure Pursuit for Robot Path Tracking},
  journal = {Autonomous Robots},
  volume  = {47},
  pages   = {685--694},
  year    = {2023},
  url     = {https://arxiv.org/abs/2305.20026}
}

@article{open_kin_planner,
  author        = {Macenski, S. and Booker, M. and Wallace, J.},
  title         = {Open-Source, Cost-Aware Kinematically Feasible Planning for Mobile and Surface Robotics},
  journal       = {arXiv preprint},
  archivePrefix = {arXiv},
  eprint        = {2401.13078},
  year          = {2024},
  url           = {https://arxiv.org/abs/2401.13078}
}

@article{temmink_wetlands,
  author  = {Temmink, R. J. M. and Lamers, L. P. M. and Angelini, C. and others},
  title   = {Recovering Wetland Biogeomorphic Feedbacks to Restore the World's Biotic Carbon Hotspots},
  journal = {Science},
  volume  = {376},
  number  = {6593},
  pages   = {eabn1479},
  year    = {2022},
  doi     = {10.1126/science.abn1479}
}

@article{mobile_robot_positioning,
  author  = {Semborski, J. and Idzkowski, A.},
  title   = {A Review on Positioning Techniques of Mobile Robots},
  journal = {Robotic Systems and Applications},
  volume  = {4},
  number  = {1},
  pages   = {30--43},
  year    = {2024},
  doi     = {10.21595/rsa.2024.23893}
}

@article{wang_offroad_planning,
  author  = {Wang, N. and Li, X. and Zhang, K. and Wang, J. and Xie, D.},
  title   = {A Survey on Path Planning for Autonomous Ground Vehicles in Unstructured Environments},
  journal = {Machines},
  volume  = {12},
  number  = {1},
  pages   = {31},
  year    = {2024},
  doi     = {10.3390/machines12010031}
}

@article{boats,
  author  = {Dunbabin, M. and Grinham, A.},
  title   = {Quantifying Spatiotemporal Greenhouse-Gas Emissions Using Autonomous Surface Vehicles},
  journal = {Journal of Field Robotics},
  volume  = {34},
  year    = {2016},
  doi     = {10.1002/rob.21665}
}

@incollection{ekf,
  author    = {Moore, T. and Stouch, D.},
  title     = {A Generalized Extended Kalman Filter Implementation for the Robot Operating System},
  booktitle = {Advances in Intelligent Systems and Computing},
  volume    = {302},
  pages     = {335--348},
  year      = {2016},
  doi       = {10.1007/978-3-319-08338-4_25}
}

@online{phidgetspatial,
  author       = {Phidgets Inc.},
  title        = {PhidgetSpatial Precision 3/3/3},
  year         = {2024},
  url          = {https://www.phidgets.com/?prodid=1205},
  note         = {Accessed 2025-07-31}
}

@online{ardusimple_rtk2b_heading_kit,
  author       = {ArduSimple},
  title        = {ArduSimpleRTK2B Heading Basic Starter Kit IP67},
  year         = {2024},
  url          = {https://www.ardusimple.com/product/simplertk2b-heading-basic-starter-kit-ip67/},
  note         = {Accessed 2025-07-31}
}

@Article{slippage,
AUTHOR = {Strawa, Natalia and Ignatyev, Dmitry I. and Zolotas, Argyrios C. and Tsourdos, Antonios},
TITLE = {On-Line Learning and Updating Unmanned Tracked Vehicle Dynamics},
JOURNAL = {Electronics},
VOLUME = {10},
YEAR = {2021},
NUMBER = {2},
ARTICLE-NUMBER = {187},
URL = {https://www.mdpi.com/2079-9292/10/2/187},

}

\end{document}